\documentclass[11pt,a4paper]{article}
\usepackage{times,latexsym}
\usepackage{url}
\usepackage[T1]{fontenc}
\usepackage[acceptedWithA]{tacl2021v1}

\usepackage[capitalise]{cleveref}

\def\Snospace~{\S{}} 

\usepackage{booktabs}
\usepackage{adjustbox}
\usepackage{colortbl}
\usepackage{soul}
\usepackage{amssymb}
\usepackage{multirow}
\usepackage{stfloats}

\usepackage{xspace}

\newcommand{\conf}{linguistic confidence\xspace}
\newcommand{\Conf}{Linguistic confidence\xspace}
\newcommand{\lingcal}{linguistic calibration\xspace}
\newcommand{\lingcaled}{linguistically calibrated\xspace}
\newcommand{\probcal}{probabilistic calibration\xspace}
\newcommand{\pipeline}{calibrator-controlled chatbot\xspace}
\newcommand{\validset}{the \textsc{Valid Set}\xspace}
\newcommand{\trainset}{the \textsc{Train Set}\xspace}
\newcommand{\testset}{the \textsc{Test Set}\xspace}
\newcommand{\intermediatemodel}{only-certainty-controlled model\xspace}  

\newlength{\extramargin}
\usepackage{calc}

\title{Reducing conversational agents' overconfidence\\ through linguistic calibration}

\author{Sabrina J. Mielke$^{1,2}$ \hspace*{.3em} Arthur Szlam$^2$ \hspace*{.3em} Emily Dinan$^2$ \hspace*{.3em} Y-Lan Boureau$^2$ \\
    ${}^1$ Department of Computer Science, Johns Hopkins University \hspace{.18in}
    ${}^2$ Facebook AI Research \\
    \texttt{sjmielke@jhu.edu} \hspace*{.3em} \texttt{\{aszlam,edinan,ylan\}@fb.com} \\}

\date{}

\begin{document}

\pagestyle{plain}
\thispagestyle{plain}

\maketitle

\begin{abstract}
While improving neural dialogue agents' factual accuracy is the object of much research, another important aspect of communication, less studied in the setting of neural dialogue, is transparency about ignorance.
In this work, we analyze to what extent state-of-the-art chit-chat models are {\it linguistically calibrated} in the sense that their verbalized expression of doubt (or confidence) matches the likelihood that the model's responses are factually incorrect (or correct).  We find that these models are poorly calibrated, yet we show that likelihood of correctness can accurately be predicted.
By incorporating such metacognitive features into the training of a controllable generation model, we obtain a dialogue agent with greatly improved linguistic calibration.

\end{abstract}

\setcounter{footnote}{0}

\section{Introduction}

\begin{figure}[t]
    \centering
    \hspace*{-.5em}
    \includegraphics[width=.9\linewidth]{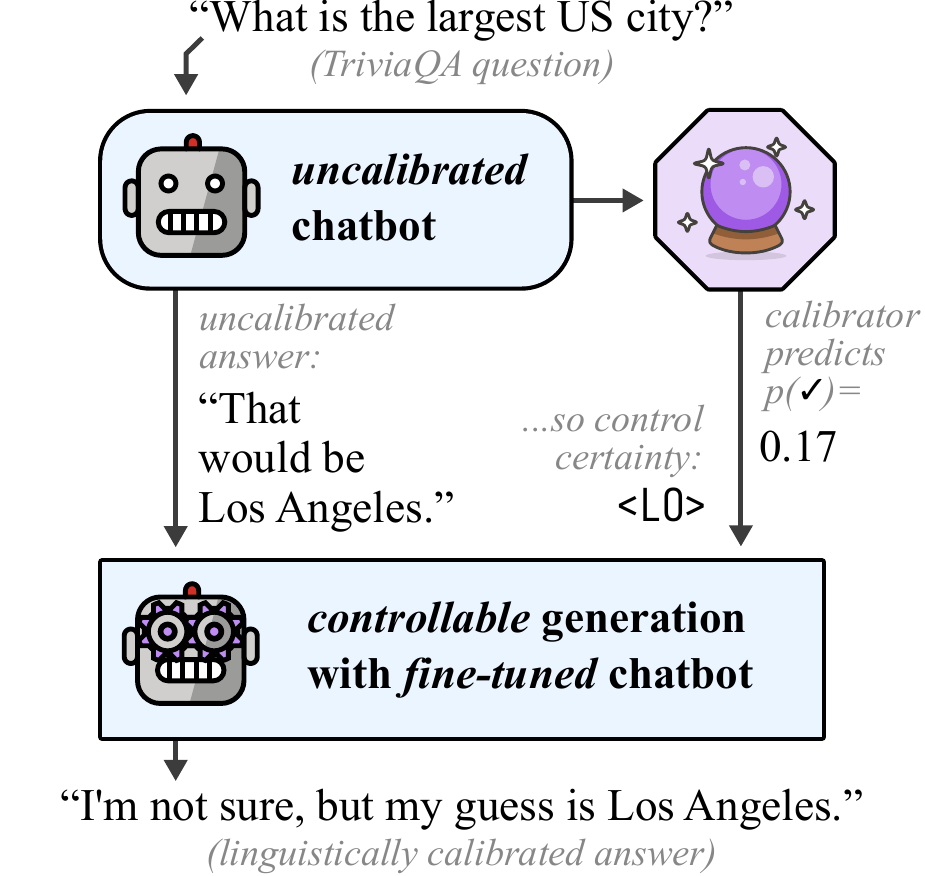}
    \caption{
        \textbf{Proposed method for re-calibrating a generative dialogue agent.} This pipeline involves a calibrator which returns the probability that the original dialogue agent's answers are correct, as well as a fine-tuned model which controls for \conf; the \conf is adjusted based on the probability returned by the calibrator, yielding a response for which the \conf aligns with the likelihood that the dialogue agent's answer is correct. This is our proposed \pipeline.
    }
    \label{fig:pipelinemodel}
\end{figure}

Neural generative open-domain English-language dialogue agents have made progress towards the ability to carry on chit-chat conversations with humans \citep{adiwardana2020meena,roller2020recipes}. Recent models---trained on large swaths of data from the internet to mimic human-human conversations---can name their favorite sports teams, describe what it's like to be the owner of two dogs, or share their opinions on tacos. However, ask a state-of-the-art chatbot ``\emph{Which is heavier, 1 kg feathers or 1 kg stone?}'', and it might confidently answer: ``\emph{Feathers, because they are heavier than a kilogram of any other material.}''\footnote{Answer generated by BST 2.7B \citep{roller2020recipes}.}
This amusing overconfidence can become problematic if someone genuinely doesn't know the answer and is misled into believing something false. Generative chit-chat dialogue agents have many issues going much beyond inaccurate answers \citep{xu2020recipes,bender2021dangers}, making them currently generally unsuitable for applications other than entertainement and research. Nevertheless, better control of the alignment between the confidence of an answer and its likelihood of being correct 
seems like a promising type of remediation: it makes models more transparent about their limitations \emph{directly in the dialogue} rather than through extrinsic instructions for adequate use that people might overlook or forget.
This goal applies Grice's maxim of quality \citep{grice1975logic} on a metacognitive level, i.e., being truthful about what one knows.
Here, this would mean that if we can train accurate predictors of correctness from information available to the model (input words and internal representations), then model generations should convey that information.
The skill of handling uncertainty would be desirable even if accuracy on factual questions ever became perfect: some questions do not have known answers, or have answers which depend on a context that a dialogue agent cannot know, making it perilous to ``ignore ignorance" \citep{smithson2012ignorance,ravetz1993sin}. 

In this work, we seek to understand whether a model's verbalized expression of confidence (``\emph{Obviously, ...}'') or doubt (``\emph{I'm not sure, but...}'') in its answer---which we refer to throughout as \emph{\conf}---corresponds to the likelihood that the answer is correct, and if not, whether we can fine-tune the models with controlled generation techniques to achieve better alignment. In other words, do state-of-the-art open domain dialogue agents ``know'' what they do not know? If yes, can this knowledge inform their responses, to achieve better verbalized metacognition? 

We thus make three main contributions. (1) We annotate a state-of-the-art chit-chat model's responses to a large-scale QA task for both factual correctness and \conf.\footnote{This data is released through the ParlAI framework at \url{https://parl.ai/projects/metacognition/}.} (2) Using these annotations, we find that the model is poorly calibrated, in that \conf does not match factual correctness, but we show that we can train a much better correctness predictor directly from the chit-chat model's representations. 
(3) We use this trained predictor within a controllable generation model to create a pipeline which greatly improves the calibration of a state-of-the-art chit-chat model.

\section{Related Work}

\paragraph{Knowledge in Open-Domain Chatbots} 
We focus on neural generative open-domain dialogue agents,
rather than general purpose language models or QA models trained to produce a factual answer given a question.
Much progress has been made
by training large-scale Transformer \citep{vaswani2017attention}  encoder-decoder models for dialogue tasks \citep{roller2020recipes,adiwardana2020meena,zhang2019dialogpt}. These sequence-to-sequence models are typically trained on large amounts of data from the internet to produce a conversational response given a dialogue history as input. Despite impressive performance on chit-chat tasks, these models are often prone to hallucinating knowledge \citep{roller2020recipes}.  \citet{dinan2018wizard} and \citet{gopalakrishnantopicchat2019} have proposed additional conditioning on a knowledge base to address this issue, but success is only partial, so we are far from being able to assume that even a knowledge-conditioned model reliably gives correct answers.

\paragraph{Overconfidence} 

  Humans' assessments of their own accuracy (\emph{confidence}) routinely exceed their objective accuracy (\emph{correctness}) \citep{pallieroverconfidence}.
  This \emph{overconfidence effect} has been well-established,
 robustly showing that humans are poorly \emph{calibrated} when completing general knowledge tasks \citep{Juslin1994TheOP,KleitmanStankovOverconfidence,Stankov1996,Stankov1998}. 
\citet{kamath-etal-2020-selective} attempt to correct overconfidence in neural
models, by 
 training QA models to abstain from answering questions in which they are likely to err, using \probcal (see next paragraph). We instead focus on getting conversational models to communicate their confidence verbally,
 i.e., still produce an answer, but one
 less misleading as to its expected correctness.

\paragraph{Probabilistic Calibration}
Much work has been dedicated to
the \probcal of deep neural networks.
\citet{pmlr-v70-guo17a} show that modern neural networks for classification tasks are poorly calibrated:
models' confidence estimate that their answer is correct doesn't match the empirical rate of correctness.
This contrasts with previous findings that show that (earlier) neural networks are well-calibrated on binary classification tasks \citep{niculescu2005}. 
We thereafter refer to this notion of calibration as \emph{\probcal} to distinguish it from \emph{\lingcal}.
More recently,  \probcal
has been explored in the space of large-scale language models (LMs). \citet{desai-durrett-2020-calibration} find that the pre-trained Transformers RoBERTa \citep{liu2019roberta} and BERT \citep{devlin2019bert} are well-calibrated in-domain on the tasks of Natural Language Inference (NLI), paraphrase detection, and commonsense reasoning. Similarly, \citet{jagannatha2020} calibrate BERT and DistilBERT \citep{sanh2019distilbert} for Part-of-Speech tagging (POS), Named Entity Recognition (NER), and QA tasks. Rather than using LMs as target predictors on classification tasks like NLI and NER, \citet{jiang2020know} instead focus on LMs as natural language generators and analyze T5 \citep{t52020}, a large scale Transformer with an encoder-decoder architecture. The authors find that it is poorly calibrated in its probability estimates on QA tasks. Conversely, \citet{radford2019gpt2} find that GPT2 is reasonably well calibrated on QA tasks, with an accuracy of 63.1\% on the 1\% of questions it is most confident in on Natural Questions \citep{naturalquestions2019}.

\paragraph{Controlled Response Generation}
We aim to reformulate answers while controlling for their expressed certainty. This requires style transfer or controlled generation techniques, which encourage certain attributes to fit prescribed values, for example a given length or sentiment. \citet{lample2018multiple} proposed a method to exert simultaneous control over multiple attributes based on concatenated learned control tokens. We similarly condition on an initial source text and concatenate multiple control tokens when generating responses.
\citet{keskar2019ctrl} trained a large-scale language model with control codes that govern style, content, and task-specific behavior. 
In the context of open-domain dialogue, \citet{see2019goodconversation} used control on attributes such as number of questions with the aim of  maximizing engagingness of dialogue models. Using larger
state-of-the-art conversational architectures,
 \citet{smith2020controlling} and \citet{madotto2020plug} compared several methods to achieve control in conversation; here, we use the simple method of training attribute-specific control tokens that was the most effective in \citet{smith2020controlling} for a variety of styles.
While our experiments in \autoref{sec:results:calibrator} suggest that good correctness prediction performance can be achieved using just the question without yet committing to the substance of an answer, which would make less constrained text generation useful, the initial goal of this paper is to control the \conf of an answer without changing its substance. For this, techniques that condition on a source response are more relevant to us than less tightly constrained controlled techniques. 
Retrieve-and-refine generation \citep{weston2018retrieve,roller2020recipes} conditions on a possible answer, but does not control the style of the response. Here, we condition on the initial answer produced
by a vanilla conversational model rather than a retrieval model, and then add additional control tokens to control the style.

\section{Quantifying Linguistic Confidence}

\begin{figure}[t]
    \centering
    \begin{adjustbox}{width=\linewidth}
        \begin{tabular}{rrl}
            \multicolumn{3}{l}{\em Axis: \conf} \\[.2em]
            & \hspace*{-.6em}\texttt{DK} &
            none: admits not to know \\
            & \hspace*{-.6em}\texttt{LO} &
            low: expresses uncertainty \\
            & \hspace*{-.6em}\texttt{HI} &
            high: confidently answers \\[1em]
            \multicolumn{3}{l}{\em Axis: correctness} \\[.2em]
            \multirow{2}{*}{\hspace*{0em}\raisebox{-.1em}{\includegraphics[width=.6em]{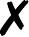}}}
            & \hspace*{-.6em}\texttt{OTHER} &
            absurd/unrelated/no answer \\
            & \hspace*{-.6em}\texttt{WRONG} &
            incorrect but not absurd answer \\
            \multirow{2}{*}{\hspace*{0em}\includegraphics[width=.65em]{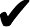}}
            & \hspace*{-.6em}\texttt{EXTRA} &
            correct, but adds incorrect knowledge \\
            & \hspace*{-.6em}\texttt{RIGHT} &
            correct and no incorrect additions \\[1em]
            \multicolumn{3}{l}{\em Not classifiable:} \\[.2em]
            & \hspace*{-.6em}\texttt{OT} &
            completely ignores the question \\
        \end{tabular}
    \end{adjustbox}
    \caption{
        \textbf{A taxonomy of \conf and correctness} for TriviaQA answers provided by a dialogue agent, yielding $3 \times 4 + 1=13$ classes.
    }
    \label{fig:taxonomy}
\end{figure}

\paragraph{Linguistic Confidence}

We aim to align a model's expressed confidence with its actual correctness, rather than increase that correctness.
We focus on models' \conf, i.e., determined by its linguistic choices (e.g. ``\emph{I don't know, but...}'' vs. ``\emph{Obviously, it's...}'').
Do these models' responses reflect whether they ``know" what they do not know (\emph{metacognition})? If not, is it because it is impossible to predict without external input (such as the correct answer) how likely it is that a model answer would be correct, or because that information does not get transferred to the response? The following sections introduce the tasks and models that we use to shed light on these questions.

\paragraph{Closed-book QA as a testbed}

The task of Question Answering (QA) traditionally has a model answer a general factoid question that a user might ask, allowing the model to consult given supporting evidence, e.g., search results or related Wikipedia articles, to give an answer.%
\footnote{Sometimes, the task of Reading Comprehension is also referred to as QA, but there, models are given specific paragraphs of texts and asked to answer questions \emph{about} that paragraph \emph{using} that paragraph.}

In this work, models do not have access to supporting evidence. Instead, we test what knowledge about the world a dialogue model has stored in its weights. Forcing a model to generate thus is called \emph{closed-book QA} \citep{t52020}, and any factoid-style question answering dataset can be used in this manner. Following GPT-3 \citep{brown2020language}, we use TriviaQA \citep{joshi-etal-2017-triviaqa} as our dataset, as it covers a large output space (unlike WebQuestions \citep{webquestions2013berant}, which is restricted to Freebase), and contains fully grammatical questions as opposed to search queries (unlike Natural Questions \citep{naturalquestions2019}, which contains ungrammatical search queries).%

To convert it into a closed-book QA dataset we can use, we merge the dataset's ``Web'' and ``Wikipedia'' sections (including shared questions only once), remove all provided evidence documents for the questions, strip the (Wikipedia-based) aliases of their `` (disambiguation)'' suffix, and then use these aliases to create a list of allowable gold answers.
We end up with 76523 question-answer pairs in the training set and 9961 in the validation set.
An example entry in this dataset looks like this:

\begin{quote}
What is the name of the tool used to sharpen a knife? \emph{(Steel, Crude steel, Long steel products, Steel, Steel (alloy), Steel (metal), Steel Construction, Steel in Africa, Steel industry, Steel manufacture, Steel plate, Steel sheeting, Steel truss, Steel worker, Steel workers, Steels, Steelworker, Steelworkers, Titanic steel, Unwrapped steel)}
\end{quote}

Despite the list of aliases of the gold answer (``Steel,'' given first in the otherwise alphabetically sorted list), evaluating correctness of answers may not always be so straightforward---consider this example answer:%
\footnote{This answer was generated by the vanilla BST 2.7B model we consider in \autoref{sec:models}, and shows that human annotations are not always reliable:
all three annotators judge the certainty of this response to be \texttt{LO}, even though the answer itself expresses no doubt. As for correctness, two say \texttt{WRONG} and one says \texttt{CORRECT}, reflecting uncertainty as to how a factually correct answer not included in the allowable gold answers should be graded.}
\emph{``It is called a whetstone.  It is a stone that is used for sharpening knives.''}

\paragraph{Annotation scheme}\label{sec:annotation_scheme}

\begin{figure*}[tbh]
    \small
    \centering
    \includegraphics[width=\linewidth]{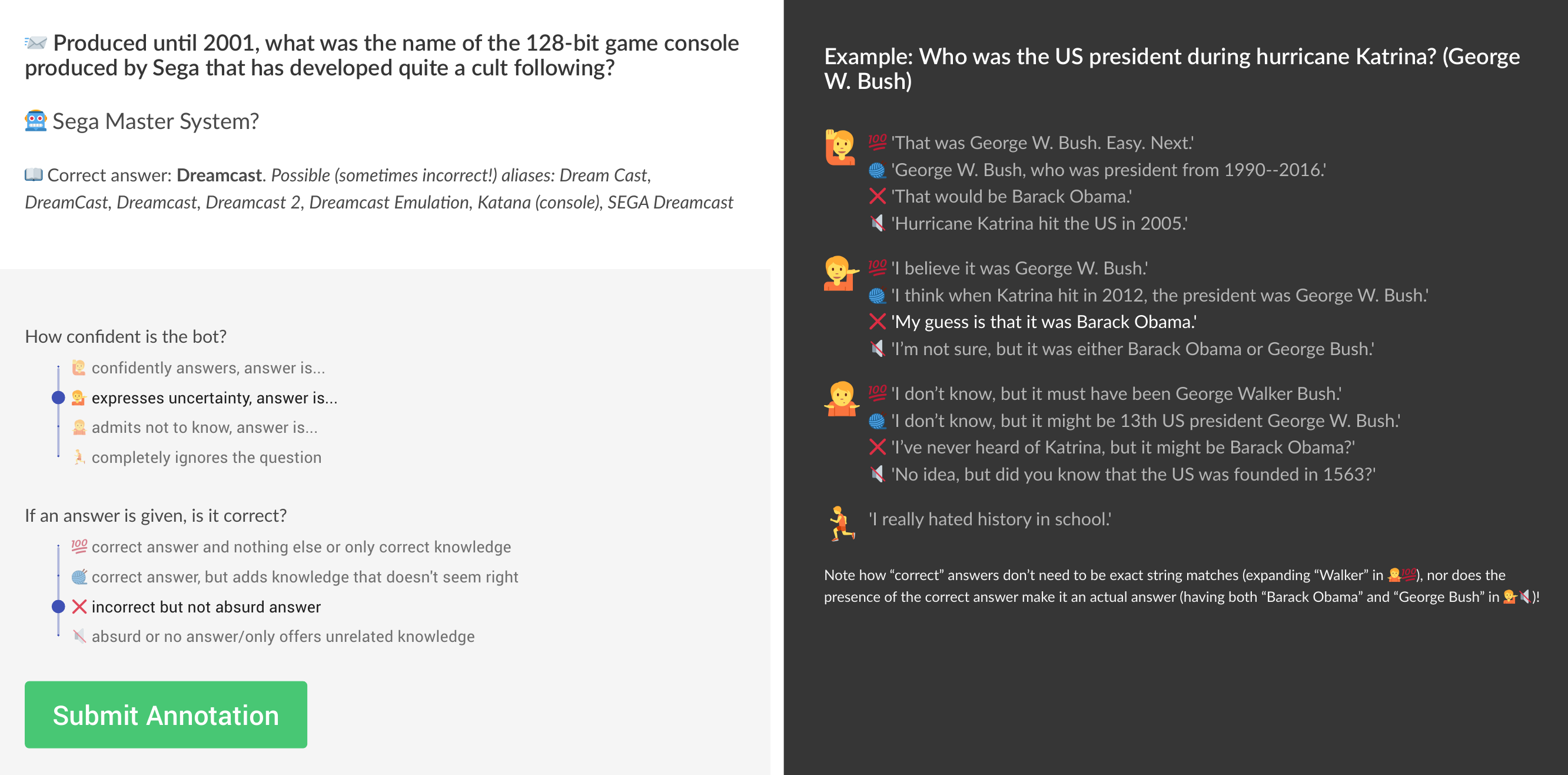}
    \caption{Human-written example answers to the question ``Who was the US president during hurricane Katrina?'' (correct answer: George W. Bush), annotated for both \conf and correctness, using the taxonomy given in \autoref{fig:taxonomy}. Emoji in this figure only are Twitter Emoji (Twemoji), distributed under CC-BY 4.0.}
    \label{fig:bush-example}
\end{figure*}

The answers that a chatbot gives for a question are full-length sentences that may or may not answer the question, may or may not do so correctly, and may or may not express confidence linguistically.
We settle on relating such generations to the gold answer aliases in our dataset by having humans annotate generations according to the annotation scheme shown in \autoref{fig:taxonomy}.  Unless the question is not even acknowledged as such (\texttt{OT}, short for ``off-topic''), the chatbot's response is judged for \emph{\conf} and for \emph{correctness} with respect to the provided gold answers.
\autoref{fig:bush-example} illustrates all 13 resulting classes with example answers in the GUI that is presented to human annotators.

The fine-grained 4-way splitting of correctness is designed to provide guidance to human annotators and reduce ambiguity. After the initial annotation, we simplify all correctness annotations to \emph{binary} correctness that better aligns with the type of linguistic framing we would like the model to be able to express, mapping \texttt{OTHER} and \texttt{WRONG} to \emph{incorrect} (\raisebox{-.1em}{\includegraphics[width=.6em]{imgs/cross}}) and \texttt{EXTRA} and \texttt{RIGHT} to \emph{correct} (\includegraphics[width=.65em]{imgs/check}).

The 3-way splitting of confidence is intuitively richer than simply splitting along confident vs. not confident (\texttt{HI} vs. not), however many responses were of the kind "I don't know, but I know that...," which makes them ambiguous. Note that the minimum length of responses enforced by the model rated as most engaging in \citet{roller2020recipes} precludes responding with a straight "I don't know," which likely makes the ambiguity more salient (see discussion of minimum length in \autoref{sec:models}). We nevertheless release the full 3-way annotations in case they are useful for further research. 

\paragraph{Automatic annotation}\label{sec:auto_annotation}

\begin{figure*}[t]
    \centering
    \hspace*{.5em}
    \begin{minipage}{.5\linewidth}
        \small
        \begin{tabular}{rrrrrrr}
                & \multicolumn{6}{c}{\em human-annotated correctness of bot answers} \\
                \cmidrule(lr){2-7}
                & \multicolumn{4}{c}{\em 4-way} & \multicolumn{2}{c}{\em binary} \\
                \cmidrule(lr){2-5} \cmidrule(lr){6-7} 
            \smash{\rotatebox{90}{\parbox{4em}{\bf gold in\\answer?}}} &
                \rotatebox{90}{
                \texttt{OTHER}} & 
                \rotatebox{90}{
                \texttt{WRONG}} & 
                \rotatebox{90}{
                \texttt{EXTRA}} & 
                \rotatebox{90}{
                \texttt{RIGHT}} &
                \raisebox{.4em}{\includegraphics{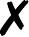}} & \raisebox{.4em}{\includegraphics{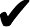}} \\[.1em]
            \toprule
            $\not\in$ & 72.95 & 20.40 & 0.17 & 0.35 & 93.35 & 0.52 \\[.1em]
            $\in$     &  0.69 &  0.23 & 1.73 & 3.47 &  0.92 & 5.20 \\
        \end{tabular}
    \end{minipage}
    \hfill
    \begin{minipage}{.4\linewidth}
        \small
        \setlength{\tabcolsep}{.5em}
        \begin{tabular}{rrrrrr}
            & & \multicolumn{4}{c}{\em human-annotated} \\[.5em]
            &
            & \rotatebox{90}{
            \texttt{OT}}
            & \rotatebox{90}{
            \texttt{DK}}
            & \rotatebox{90}{
            \texttt{LO}}
            & \rotatebox{90}{
            \texttt{HI}} \\[.1em]
            \toprule
            \multirow{4}{*}{\rotatebox{90}{\em\hspace*{-2em} BERT-based\hspace*{-1em}}}
            & \texttt{OT}
            & 0.56 & 0.28 & 0.11 & 0.67 \\[.1em]
            & \texttt{DK}
            & 0.17 & 7.81 & 7.81 & 0.06 \\[.1em]
            & \texttt{LO}
            & 0.06 & 14.67 & 34.63 & 0.11 \\[.1em]
            & \texttt{HI}
            & 2.73 & 0.11 & 0.33 & 29.89 \\[.1em]
        \end{tabular}
    \end{minipage}
    \caption{
        Composition of the vanilla bot's answers on the \validset (in \% of total): comparing match-based correctness scoring to human annotations (left; treating binarized human labels as gold, the match-based correctness labels have 0.85 precision and 0.91 recall) and BERT-based \conf scoring to human annotations (right; binarizing \conf into \texttt{HI} and \texttt{not-HI}, the classifier has 0.90 precision and 0.97 recall for detecting \conf).
    }
    \label{tab:match-and-bert-vs-human}
\end{figure*}

Noting predictability in patterns of human annotation, we seek to quantify whether automatic annotation would be an adequate substitute. The left half of \autoref{tab:match-and-bert-vs-human} indeed confirms that the simplified binary correctness annotations are highly predictable by simply checking whether any of the answer aliases appear in the generation (tokenized).
We will refer to this way of scoring correctness as \emph{match-based}, and use it as an automatic proxy for human annotations, when the latter is cost-prohibitive.

\Conf is harder to automatically infer using template- and match-based methods, as there are many ways to express doubt or confidence.
Still, we find that we obtain usable predictions by training a BERT-based classifier on a set of 2000 annotated question-prediction pairs.%
\footnote{These samples come from \trainset (see \autoref{sec:data_collect_results}); the classifier is the \texttt{bert\_classifier}
from ParlAI \citep{miller2017parlai}, fine-tuning the final layer and predicting output classes from the \texttt{[CLS]} token. We did not tune this model heavily, or try other tricks like averaging embeddings, as we were satisfied with performance.}
We will refer to this way of classifying 4-way certainty (\texttt{DK}, \texttt{LO}, \texttt{HI}, and \texttt{OT}) as \emph{BERT-based} and likewise use it extensively for training.
This classifier works well (see the right half of \autoref{tab:match-and-bert-vs-human}) for distinguishing \texttt{DK}/\texttt{LO} from \texttt{HI}, but struggles to discern between \texttt{DK} and \texttt{LO} (likely due to inconsistency in human annotation for this distinction, as noted above), and to a lesser degree \texttt{OT} and \texttt{HI}.

\paragraph{Models}\label{sec:models}

Our base model is the state-of-the-art open-domain English-language dialogue system BlenderBot from \citet{roller2020recipes}. ``BlenderBot'' refers to a suite of models of varying sizes which employ a Seq2Seq Transformer architecture \citep{vaswani2017attention}. These models were pretrained  on 1.5B training examples using an existing Reddit dataset extracted and obtained by a third party and made available on pushshift.io \citep{baumgartner2020pushshift}.\footnote{\url{https://files.pushshift.io/reddit/}}
We use the 2.7B parameter version that is finetuned on the Blended Skill Talk tasks \citep[BST;][]{smith2020bst} and consider the outputs of beam search using the models' recommended standard parameters, which includes a requirement for generated answers to have at least 20 tokens.
We choose this model (referred to as ``vanilla'' from here on) because it is the configuration that is rated as most engaging by humans \citep{roller2020recipes} and therefore the most realistic use-case, even though it is not the best-performing QA model.%
\footnote{
    It is worth noting that removing the minimum length requirement and not fine-tuning on BST did improve QA performance slightly (from 5.0\% to 6.9\% accuracy on \validset), and increasing the model capacity to 9.4B parameters even raised it to 8.5\% accuracy. Improving model capacity without suffering losses in engagingness is an important avenue for further research that is orthogonal to our proposal.
}
This vanilla model attains an accuracy of only 4.8\% on the test set,%
\footnote{
    We also experimented with top-$k$ and nucleus sampling, which slightly reduced accuracies, and looked at correctnesses of the top few beams instead of just the single most likely generation, but those usually were similar to the top-1 answer in terms of correctness.
    }
yet it answers 29.45\% of questions confidently (\texttt{HI}), making only 14\% of the model's confident answers actually correct (see \autoref{tab:vanilla-is-poorly-calibrated}).

We also try to examine what kind of questions are intrinsically ``difficult'' in a way that can be detected by shallow features. For example, we might hypothesize that questions about locations might be easier than questions about people---this would be reflected by the words ``where'' and ``who'' in a question being predictive of correctness.
To obtain such predictive surface features we train a single sparse logistic regression model on all 2, 3, \ldots, 7-grams that appear at least 5 times in our human-annotated test set to predict binarized correctness and binarized certainty from questions (1166 such $n$-grams) or from answers (1882 such $n$-grams). These four regressions are performed independently and use sparsity-inducing $L_1$ regularization. 
This yields between 9 and 19 $n$-grams that are useful indicators, the three most negative and positive are shown in \autoref{tab:ngrams}.

\begin{table}[tbh]
    \centering
    \setlength{\tabcolsep}{.4em}
    \begin{tabular}{c|rl|rl}
        \toprule
        \multicolumn{5}{l}{\bf Correctness} \\
        & \multicolumn{2}{l}{\bf from questions} & \multicolumn{2}{l}{\bf from answers} \\
        \midrule
        & 1.098 & city is & 0.506 & It is the \\
        \smash{\rotatebox{90}{$\to$}} & 0.187 & $\gg$ What & 0.502 & It was a \\
        \smash{\includegraphics{imgs/check}} & 0.155 & is the & 0.375 & used to \\
        \midrule
        \smash{\raisebox{-.1em}{\includegraphics{imgs/cross}}} & -0.292 & $\gg$ What was & -0.595 & I do \\
        \smash{\raisebox{-.1em}{\rotatebox{90}{$\leftarrow$}}} & -0.658 & $\gg$ Which & -0.685 & but I \\
        & -0.792 & $\gg$ Who & -0.874 & I don't \\
        \bottomrule
        \toprule
        \multicolumn{5}{l}{\bf Certainty (\texttt{OT}/\texttt{DK}/\texttt{LO} $\le$ \texttt{HI})} \\
        & \multicolumn{2}{l}{\bf from questions} & \multicolumn{2}{l}{\bf from answers} \\
        \midrule
        & 0.737 & is a & 0.812 & $\gg$ It \\
        \smash{\rotatebox{90}{$\to$}} & 0.565 & in which & 0.152 & in the \\
        \texttt{HI} & 0.193 & is the & 0.005 & $\gg$ The \\
        \midrule
        \texttt{LO} & -0.355 & in the & -2.459 & $\gg$ I \\
        \texttt{DK} & -0.540 & $\gg$ Who & -2.750 & but I \\
        \texttt{OT} & -0.782 & $\gg$ Which & -4.122 & I'm not \\
        \bottomrule
    \end{tabular}
    \caption{
        Predictive $n$-grams (with $n \in \{2, \ldots, 7\}$) in questions and answers with their associated weights, negative weights indicating a push towards ``correct'' and \texttt{OT}/\texttt{DK}/\texttt{LO}, and positive weights counting towards ``incorrect'' and \texttt{HI}.
    }
    \label{tab:ngrams}
\end{table}

\section{Re-calibrating chatbots' language}

Given that BST 2.7B and all other BlenderBot variants are poorly \lingcaled (specifically, overconfident in answers to TriviaQA questions), 
we introduce a pipeline for improving calibration.

\paragraph{Pipeline overview}
We propose training a calibrator and using controllable generation techniques to
allow generative dialogue agents to better ``own their ignorance,'' i.e.,
such that the models' \conf
better aligns with the probability that the answers are correct. 
The overall pipeline is illustrated%
\footnote{The robot emoji in this figure was drawn by Mariella Steeb and distributed as part of the OpenMoji project under CC-BY-SA 4.0. The crystal ball illustration was drawn by Vincent Le Moign and is distributed as part of the Streamline Emoji Project under CC-BY 4.0.}
in \autoref{fig:pipelinemodel}. We first train a calibrator to return the empirical probability that the model's answer is correct (without seeing the gold answer),
and finetune the generative dialogue model to enable control over \conf.
Using the calibrator and the controllable generation model, we adjust the dialogue agent's response by choosing \conf control tokens that align with the probability returned by the calibrator, resulting in a \emph{\pipeline}.

\paragraph{Training a calibrator}\label{sec:methods_training_calibrator}

The first step
involves training a calibrator that predicts the probability that the model's response is correct, given the question and answer,
and the vanilla model's internal representations
of both.
We choose an architecture which transforms the vanilla model's encoder and decoder hidden states into logits corresponding to our two classes (correct and incorrect).%
\footnote{The model applies a linear layer followed by GELU activation \citep{hendrycks2016gelu} to all states individually, aggregates the resulting vectors via a max pooling operation, and finally, transforms that result using a linear-GELU-linear MLP to return logits. All hidden layers are of size 256.}
The model is trained using 50,000 questions from the full TriviaQA training split with the vanilla model's corresponding responses, automatically annotated for correctness using the match-based annotation scheme (see \autoref{sec:auto_annotation}). Ablations in \autoref{sec:results:calibrator} show that different models for the calibrator, some not using the answer, some not using the internal representations, yield similar results.

\paragraph{Training a controllable generation model}\label{sec:methods_control_gen}

The next step
trains a generative model that will adjust the \conf of a response, provided the original response and a control token representing the desired \conf: \texttt{<DK>}, \texttt{<LO>}, or \texttt{<HI>}. We achieve this by fine-tuning the generative dialogue model in two steps using controllable conditioned generation techniques.

\paragraph{Stage 1: confidence controllable model} We first train a \conf controllable generative dialogue model following the method in \citet{smith2020controlling}. We fine-tune the vanilla model on the original BST tasks, augmented with an additional task constructed from TriviaQA to incorporate confidence signals: 25,000 questions from the TriviaQA training split are augmented with a \emph{control token} capturing the vanilla model response's \conf, as given by the BERT-based classifier (\autoref{sec:auto_annotation}).
The expected output is the vanilla model's response to the question. 
All incorrectly answered examples and examples with the \texttt{OT} label are discarded, and remaining examples are oversampled to have the same overall certainty distribution as we see on \validset.
The model thus learns to associate the \conf of the response with the control tokens and can generate responses with a desired degree of
confidence at inference time by setting appropriate control tokens.
We refer to this model as the \emph{\intermediatemodel}.

\paragraph{Stage 2: confidence-and-content controlled model}
Adjusting the \conf of a generated response via control tokens with the \intermediatemodel often also changes the \emph{content} of the response. 
Simultaneous control over both \conf \emph{and} content would be preferable, to allow changing the \conf of a given response without altering the provided answer for a question.
We achieve this in a second stage of fine-tuning by constructing a task that simultaneously conditions on \conf and response content. 
Training prompts for this task are constructed
by concatenating the same 25,000 TriviaQA training split questions with the vanilla model's response,
a \conf control token as before, and also an additional control token capturing whether the content of the \intermediatemodel's response when given that question and \conf control token is the same (\texttt{<SAME>}) or different (\texttt{<DIFF>}) from the vanilla model's response.
The expected output is the \intermediatemodel's response to the question with that \conf control token. 
The content control token is \texttt{<SAME>} if both the vanilla model and \intermediatemodel's responses to the question are correct, and \texttt{<DIFF>} if only one of them is correct. Examples where both the vanilla model and \intermediatemodel's responses are incorrect are discarded, because there are so many different ways to be incorrect. Choosing \texttt{<SAME>} at inference time yields a model which adjusts the \conf of the vanilla model's response (provided as input) without changing the answer to the question.
We refer to this model as our ``controlled'' model, to be used in the final pipeline.

\section{Results}\label{sec:results}

We describe data collection and annotation results, as well as experimental results and analysis on the vanilla model and each stage of the pipeline for the \pipeline.

\subsection{Data collection and annotation}\label{sec:data_collect_results}

We collect human annotation for both training data and for our final evaluation of the vanilla model and the \pipeline. Question and response pairs are annotated for both correctness and \conf using the annotation scheme described in \autoref{sec:annotation_scheme}. Crowdsource annotators annotate questions in batches of nine questions, after completing an ``onboarding'' test of three questions.

\paragraph{Training data} We collect annotations for the vanilla model's responses to 2000 questions each from the train and validation splits of TriviaQA. Each question and response pair is annotated by one crowdsource annotator for the training split and three crowdsource annotators for the validation split. We refer to these splits as \trainset and \validset throughout; we use \trainset to train the BERT-based classifier (\autoref{sec:auto_annotation}) and for early-stopping the calibrator training, we use \validset for early-stopping the controllable generation model fine-tuning steps and for tuning hyperparameters for BERT-based classifier, calibrator, and the controllable generation models.

\paragraph{Final evaluation data}
Three annotators label 5000 question and response pairs from the TriviaQA validation split (none of which overlap with \validset) for each the vanilla model and the controlled model under all three \conf control settings (\texttt{DK}, \texttt{LO}, \texttt{HI}). We refer to this size 3 $\times$ 4 $\times$ 5000 set as \testset throughout.
Note that evaluating our \pipeline would only require annotating responses generated with the one \conf control token dictated by the probability returned by the calibrator for each example. 
However, collecting annotations for all three \conf control settings allows future work to improve the calibrator in isolation, without having to re-train and re-label the controlled outputs.

\paragraph{Inter-annotator agreement} 
We analyze agreement between annotators using the question and response pairs from \validset that were annotated three times each. For \conf, 43.60\% of samples have all three annotators agree and 97.60\% have at least two agree. For four-way correctness, these ratios are 69.15\% and 97.90\%; for binary correctness, they are 94.35\% and 99.40\%.
We restrict to samples for which a majority (binary on correctness) exists and take the majority label, reducing the size of \validset from 2000 to 1793 examples and the size of \testset from 5000 to 4793 examples.

\subsection{Calibrator training results}\label{sec:results:calibrator}

\begin{figure*}[t]
    \centering
    \includegraphics[width=.93\linewidth]{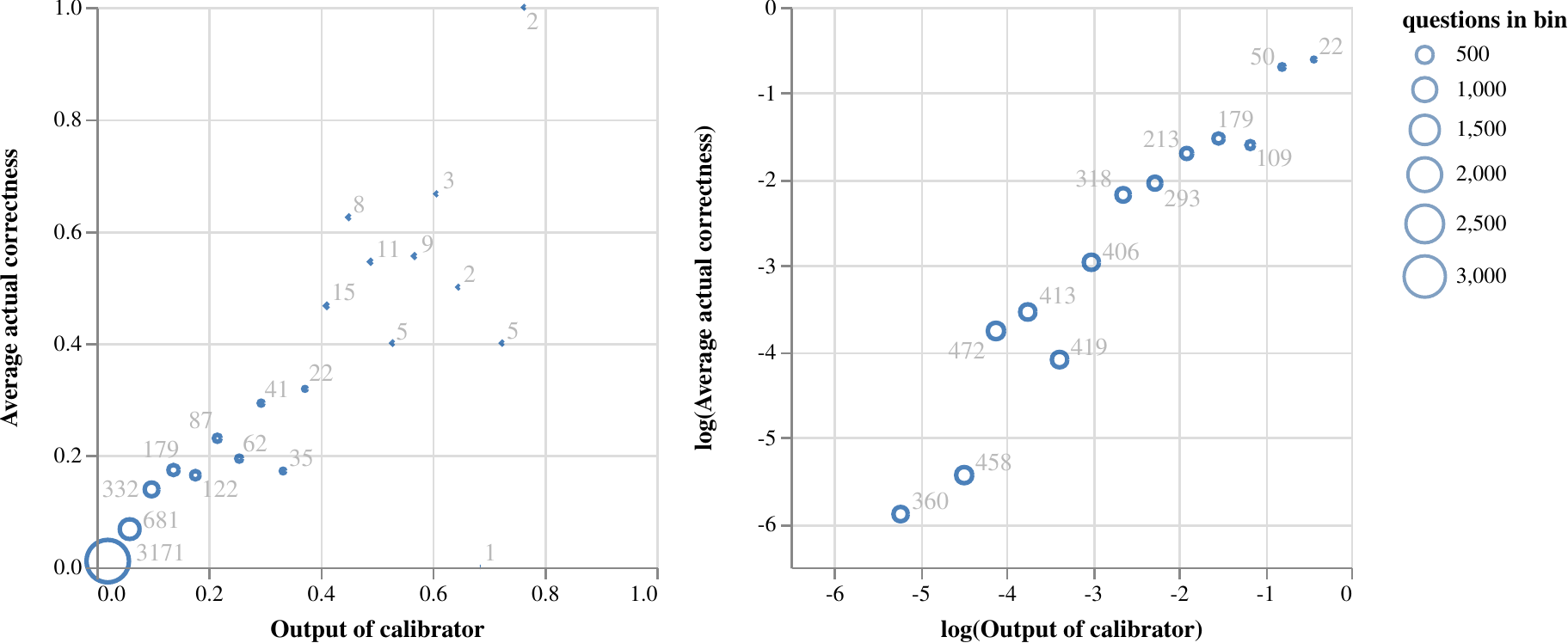}
    \caption{\textbf{Calibrator performance.}
    Performance evaluated on \testset by comparing the ratio of answers that were actually correct to the probability returned by the classifier (binned).  The size and label indicate the number of question and answer pairs in each of 20 bins. }
    \label{fig:calibrator-bubbles}
\end{figure*}

The \pipeline can only be as good as the calibrator, requiring the ability to reliably predict how likely an answer is to be correct without access to additional knowledge. 
\autoref{fig:calibrator-bubbles} plots the observed correctness on \testset against the probability predicted by the calibrator that we selected using \validset, 
and shows that the calibrator does a good job predicting correctness probability.
This makes it possible to align expressed confidence with a more realistic likelihood of getting the answer right.

We also evaluate calibration using the metrics from \citet{pmlr-v70-guo17a}. The first two metrics assume that examples are sorted into equally-spaced bins by their predicted likelihood of correctness (which thus need not contain the same number of samples).
We can define the ``distance'' between the predicted likelihood of correctness of a bin (the midpoint between the start and the end of the bin) and the \emph{actual} correctness of the bin (the average of all individual examples, counting correct ones as 1, incorrect ones as 0)---lower is better.
Using these distances, the Expected Calibration Error (ECE) refers to the weighted average of all bins' distances (weighted by how many samples out of the total were in a bin)---our calibrator achieves an ECE of 0.018.
Similarly, the Maximum Calibration Error (MCE) refers to the maximum of all bins' distances---our calibrator reaches an MCE of 0.292.
Finally, we can calculate the Average Negative Log-Likelihood (ANLL) by averaging every individual example's NLL, which for correct examples means the log of the predicted likelihood of being correct, and for incorrect answers means taking the log of the inverse event, i.e., $\log 1 - p$. The calibrator reaches an ANLL of 0.165.

Note that these metrics show and reward capturing different degrees of uncertainty and incorrectness that may not be as apparent in our main results in \autoref{sec:evaluating-the-pipeline}, as most examples are low-confidence and low-correctness.

\begin{table}[tb]
    \setlength{\tabcolsep}{6pt}
    \centering
    \begin{adjustbox}{width=\linewidth}
        \begin{tabular}{lrrrrr}
            & \multicolumn{2}{c}{\em thresh. 0.375} & \multicolumn{2}{c}{\em 20 bins} & \\
            \cmidrule(lr){2-3} 
            \cmidrule(lr){4-5} 
            calibrator & ECE & MCE & ECE & MCE & (A)NLL \\
            \toprule
            +enc +dec & .2021 & \bf.2289 & .0176 & \bf.2917 & .1650 \\
            \midrule
            -enc +dec    & .2017 & .2873 & .0145 & .7250 & \bf.1628 \\
            +enc -dec     & .2003 & .2870 & \bf .0061 & .7250  & .1802 \\
            -enc -dec & \bf.1989 & .3000  & .0113 & .6250  & .1786 \\
            \midrule
            BERT & .2063 & .3446 & .0156 & .7750  & .1635 \\
        \end{tabular}
    \end{adjustbox}
    \caption{Comparison of different calibrators via Expected Calibration Error (ECE), Maximum Calibration Error (MCE), and (Average) Negative Log Likelihood \citep{pmlr-v70-guo17a}. Closer to zero is better for all metrics.
    Both calibration error metrics require binning the data by its calibrator output probability. Threshold 0.375 means that we have only two bins, split on the threshold we end up choosing in the calibrator pipeline (\autoref{sec:evaluating-the-pipeline})---note that this threshold was picked using results from the +enc +dec set up, so was not optimized for the other set ups. Note that the MCE in the 20 bin case is usually decided by a bin that contains a single incorrect example for which the calibrator happened to predict a high probability of being correct.}
    \label{tab:calibrator-metrics}
\end{table}

We also experimented with training calibrators with more limited inputs to the calibrator, which could potentially allow for controlled generation based merely on the question, that we leave for future work.
The results of these ablations are shown in \autoref{tab:calibrator-metrics} and suggest that (1) even questions by themselves contain enough information to predict correctness almost as reliably as our full calibrator (+enc -dec), and (2) empirical correctness can even be predicted directly from words using an independent model (BERT, fine-tuned) to a reasonable accuracy.
This could be seen as corroboration of our $n$-gram findings in \cref{tab:ngrams}, meaning that certain kinds of questions, e.g., those asking for ``who'' and ``which,'' are intrinsically difficult and a fine-tuned BERT calibrator can pick up on the fact that the chatbot struggles with these kinds of questions.
Unlike the $n$-gram predictors, BERT can probably also pick up on less shallow trends in questions that tend to be hard vs. easy, explaining its surprisingly good performance.
So, while our existing set up shows that calibration can be achieved reasonably well without leveraging model internals (BERT can do reasonably well, too, despite different training data) or even full question-answer pairs (see the +enc -dec ablation), it does support us in our central objective, being able to predict how likely an answer is to be correct so that we can intervene correctly.
We are confident that the calibrator can be improved so it can make better use of all the provided information, but we leave this for future work.

For qualitative insight, \autoref{tab:calibrator-tops} shows all question/answer pairs for which the calibrator believes the answers are more likely right than wrong. 
Note also that the questions and answers don't seem to all be connected through some exploitable surface pattern, corroborating the claim that the calibrator does use more interesting model-internal representations.

\subsection{Controllable generation training results}

\begin{table*}[tbh]
    \centering
    \mbox{}
    \begin{adjustbox}{width=\linewidth}
        \begin{tabular}{rl}
            \toprule
            question & \textbf{How many Von Trapp children were there in the Sound of Music?}\\
            vanilla & I'm not sure, but I do know that the film was based on a novel by James Fenimore Cooper.\\
            \texttt{<DK>} & I'm not sure, but I do know that there were eight children in the film.\\
            \texttt{<LO>} & I'm not sure, but I do know that there were eight children in the film.\\
            \texttt{<HI>} & There were eight Von Trapp children, including Sally Hemings, Sally Field, and Johnny Depp.\\\midrule
            question & \textbf{Who was the only British medal winner at the 2010 Winter Olympics?}\\
            vanilla & That would be Michael Phelps. He was born and raised in Birmingham, England.\\
            \texttt{<DK>} & I'm not sure, but I do know that the winner of that event was Michael Phelps.\\
            \texttt{<LO>} & I'm not sure, but I do know that the winner of that event was Michael Phelps.\\
            \texttt{<HI>} & That would be Michael Phelps. He was born and raised in Birmingham, England.\\
            \bottomrule
        \end{tabular}
    \end{adjustbox}
    \caption{Contrasting answers from the vanilla model and the controlled model with all three certainty control tokens, over a non-cherry-picked (randomly sampled) list of questions.\\}
    \label{tab:controlled-examples}
\end{table*}

The final controllable model%
\footnote{All parameters are set as in the vanilla BST 2.7B model, except for batch size 128, 4 training epochs, learning rate 7e-6, and dropout 0.2 for both stages. For stage 1, the new task has weight 5.0; for stage 2 the new task has weight 9.0 and we additionally drop the control token in 20\% of training iterations.}
shows convincing separation of confident
from non-confident answers on \testset, as seen on two non-cherry-picked examples in \autoref{tab:controlled-examples}.
Combining \texttt{<DK>}- and \texttt{<LO>} categories (see discussion in \autoref{sec:annotation_scheme}),  98.79\% and 99.12\% of \texttt{<DK>}- and \texttt{<LO>}-forced are rated by humans as not belonging to the \texttt{HI} category, 
respectively, and 96.27\% of \texttt{<HI>}-forced generations are judged as \texttt{HI} by humans. Furthermore, 88.46\% of questions that the vanilla model answered correctly remain correct when letting the \texttt{<HI>}-forced model answer the same questions. By contrast, the \intermediatemodel (not conditioned on the initial answer itself) only maintains 56.81\% of correct answers as correct when conditioned on the \texttt{<HI>} token.
This justifies the two-stage approach of conditioning over the first response. In fact, 61.65\% of questions that were answered confidently and correctly by the vanilla model are given the word-for-word same answer by the \pipeline.
Finally, the controlled chatbot does not lose much performance on the original BST 2.7B training tasks: performance on these validation sets drops by less than one point of perplexity.

\pagebreak

\subsection{Evaluating the \pipeline}\label{sec:evaluating-the-pipeline}

\begin{figure*}
    \centering
    \hspace*{-1.2em}
    \begin{minipage}{.33\textwidth}
        \raggedleft
        \includegraphics[width=\linewidth]{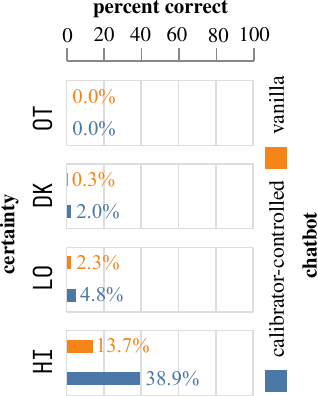}
    \end{minipage}
    \hspace*{1em}
    \begin{minipage}{.6\textwidth}
        \setlength{\tabcolsep}{.5em}
        \small
        \def\arraystretch{1.25}
        \begin{tabular}{rrrrr|rr}
            \raisebox{1.5em}{\parbox{5em}{\centering\textbf{correctness}\\\textit{(4-way and binary)}}} & \rotatebox{90}{
            \texttt{OTHER}} & \rotatebox{90}{
            \texttt{WRONG}} & \rotatebox{90}{
            \texttt{EXTRA}} & \rotatebox{90}{
            \texttt{RIGHT}} & \raisebox{.4em}{\includegraphics{imgs/cross}} & \raisebox{.4em}{\includegraphics{imgs/check}} \\
            \bottomrule
            \multicolumn{7}{l}{\rule[9pt]{0pt}{0pt}\textbf{vanilla model} (4.8\% overall accuracy)} \\
            \toprule
            \mbox{}\\[-1.5em]
            2.4\% \texttt{OT}
            & \multicolumn{4}{c|}{\color{gray} ---} & \multicolumn{2}{c}{\color{gray} ---} \\
            31.6\% \texttt{DK}
            & 96.86 & 2.96 & 0.09 & 0.09 & 99.67 & 0.33 \\
            38.1\% \texttt{LO}
            & 88.65 & 9.55 & 0.65 & 1.15 & 97.71 & 2.29 \\
            27.8\% \texttt{HI}
            & 33.80 & 54.10 & 3.60 & \cellcolor{gray!20}8.50 & 86.25 & \cellcolor{gray!20}13.75 \\
            \bottomrule
            \multicolumn{7}{l}{\rule[9pt]{0pt}{0pt}\textbf{\pipeline} (5.1\% overall accuracy)} \\
            \toprule
            \mbox{}\\[-1.5em]
            0.2\% \texttt{OT}
            & \multicolumn{4}{c|}{\color{gray} ---} & \multicolumn{2}{c}{\color{gray} ---} \\
            12.1\% \texttt{DK}
            & 75.79 & 22.40 & 0.68 & 1.13 & 98.03 & 1.97 \\
            85.9\% \texttt{LO}
            & 69.48 & 26.57 & 0.84 & 3.12 & 95.21 & 4.79 \\
            1.8\% \texttt{HI}
            & 16.92 & 47.69 & 7.69 & \cellcolor{gray!20}27.69 & 61.11 & \cellcolor{gray!20}38.89 \\
            \toprule
        \end{tabular}
    \end{minipage}
    \caption{
        Human majority annotations on the vanilla model's and the \pipeline's answers to held-out test questions, given as \% of the total for which majorities exist. Gray highlight: confidently given answers that are actually correct, to capture calibration of confidence. The plot on the left shows the average \emph{binary} correctness for both the vanilla chatbot and the \pipeline (i.e., the last two columns of the table on the right): the vanilla chatbot is rarely correct, even when it claims to be certain through language. The \pipeline has more than double the chance of being correct when it expresses certainty linguistically, compared to the vanilla model. This comes with more selective use of \texttt{HI} (and to a lesser extent \texttt{DK}), as shown on the right. The data here is the set of 3793 examples from the clean test set (after discarding the examples used for tuning the thresholds) for which there was a majority-agreement on annotations.
    }
    \label{tab:vanilla-is-poorly-calibrated}
\end{figure*}

Finally, it is time to evaluate our \pipeline and the vanilla model both on \testset, which contains 4793 examples (see \cref{sec:data_collect_results}), using full human annotations for both correctness and certainty of all evaluated models' generations.

Running the \pipeline requires mapping the empirical correctness probabilities returned by the calibrator to the control tokens used by the controllable model. For this, we select thresholds on the calibrator outputs to map to \texttt{DK}, \texttt{LO},
and \texttt{HI} by searching over all threshold values between 0 and 1 (with 0.025 steps) that maximize
$p(\includegraphics{imgs/check}\mid\mbox{{\tt HI}})$
using the first 1000 questions of \testset, which are therefore subsequently excluded from the final test set results. 
This results in thresholds of 0 and 0.375, so the calibrator is never asked to produce \texttt{DK}, even though the resulting sentence sometimes ends up being annotated as such (see also \autoref{sec:annotation_scheme} about ambiguity between both categories).

\autoref{tab:vanilla-is-poorly-calibrated} shows that our \pipeline displays much better \lingcal, with the correctness of linguistically confident answers (both judged by humans) jumping nearly threefold, from 13.7\% to 38.9\%.%
\footnote{The increase is highly significant with $p < 10^{-6}$ under a 
paired permutation test.}
Note that this is achieved by answering much fewer questions confidently, which is a necessary side effect for a chatbot for which overall correctness is low.
The full confusion matrix between vanilla and \pipeline is shown in \cref{tab:confusionmatrix}.

\begin{table}[tbp]
    \centering
    \setlength{\tabcolsep}{.5em}
    \begin{tabular}{rrrrrr}
        & \multicolumn{5}{r}{\em \pipeline} \\[.5em]
        &
        & \hspace{1em}\rotatebox{90}{\texttt{OT}}
        & \rotatebox{90}{\texttt{DK}}
        & \rotatebox{90}{\texttt{LO}}
        & \rotatebox{90}{\texttt{HI}} \\[.1em]
        \toprule
        \multirow{4}{*}{\rotatebox{90}{\em\hspace*{-2em} vanilla\hspace*{-1em}}}
        & \texttt{OT}
        & 5 & 10 & 72 & 4 \\[.1em]
        & \texttt{DK}
        & 0 & 237 & 959 & 2 \\[.1em]
        & \texttt{LO}
        & 0 & 104 & 1332 & 6 \\[.1em]
        & \texttt{HI}
        & 2 & 105 & 895 & 60 \\[.1em]
    \end{tabular}
    \caption{Confusion matrix between the vanilla chatbot's answer certainties and that of \pipeline.}
    \label{tab:confusionmatrix}
\end{table}

It is thus not surprising that just generating low-certainty responses (\texttt{LO}) also improves over the baseline, yielding a 22.2\% rate of correctness among generated answers that humans rate as highly confident (\texttt{HI}).%
\footnote{Generating with certainty \texttt{LO} yields 0.7\% \texttt{HI} answers; generating with \texttt{DK} yields 0.8\%, of which 19.4\% are correct; generating with \texttt{HI} yields 96.5\%, of which 7.9\% are correct. All these correctness rates are statistically significantly different from both the vanilla system and the \pipeline ($p < 10^{-6}$).}
Importantly, overall accuracy is not negatively impacted by our calibration procedure, but actually slightly increases from 4.8\% to 5.1\%, though this increase is not statistically significant under a paired permutation test.%
\footnote{Of the baselines described in the previous footnote, only the \texttt{HI}-forced generations that achieve an overall accuracy of 7.7\% are significantly better than the vanilla model's overall responses at $p < 10^{-6}$.}
As a further beneficial side effect, off-topic answers (\texttt{OT}) are greatly reduced in this \pipeline.
Qualitatively, 
these two cherry-picked examples illustrate how the \pipeline improves over the vanilla model in the sense that it is able to ``own its ignorance'' and verbally express doubt when its answer is likely incorrect:

\begin{quote}
\textbf{What type of animal is a frogmouth?}\\
\emph{vanilla}: Frogmouths are a type of amphisbaenia that live in freshwater.\\
\emph{\pipeline}: I'm not sure, but I think it's a type of freshwater fish.
\end{quote}

\begin{quote}
\textbf{What type of animal is a Garron?}\\
\emph{vanilla}: A garron is a type of lizard. They are native to the Americas.\\
\emph{\pipeline}: I'm not sure, but I think it's a type of lizard. It's also the name of a town in France.
\end{quote}

\begin{table*}[p]
    \centering
    \mbox{}
    \begin{adjustbox}{width=\linewidth}
        \begin{tabular}{rcp{39em}}
            prediction & label & question / answer \\\toprule
            0.782 & {\includegraphics{imgs/check}} & \textbf{What part of the body produces insulin?} Insulin is produced in the pancreas, which is located in the abdomen.\\
            0.749 & {\includegraphics{imgs/check}} & \textbf{Vishnu and Shiva are gods in which religion?} They are the gods of Hinduism, Buddhism, and Jainism.\\
            0.742 & {\includegraphics{imgs/check}} & \textbf{Las Vegas is in which US State?} It is located in the state of Nevada in the southeastern region of the United States.\\
            0.732 & {\includegraphics{imgs/cross}} & \textbf{Who is Darth Vader's son?} Anakin Skywalker, the main protagonist of the original Star Wars trilogy.\\
            0.727 & {\includegraphics{imgs/check}} & \textbf{What city do The White Stripes and Kid Rock come from?} The white stripes are from Jacksonville, Florida. Kid rock is from Detroit.\\
            0.714 & {\includegraphics{imgs/cross}} & \textbf{Which country will host the 2014 FIFA World Cup?} I think it will be hosted in Qatar, but I'm not 100\% sure.\\
            0.707 & {\includegraphics{imgs/cross}} & \textbf{Kiel and L\"{u}beck are cities in which German state?} Kiel is in Bavaria and Lubeck is in Lower Saxony.\\
            0.690 & {\includegraphics{imgs/cross}} & \textbf{Aladdin is a variety of which vegetable?} Yes, it is a type of cabbage. It is one of my favorite vegetables.\\
            0.665 & {\includegraphics{imgs/check}} & \textbf{What does Mexican Coke have that American Coke doesn't?} Mexican coke uses cane sugar instead of high fructose corn syrup.\\
            0.635 & {\includegraphics{imgs/cross}} & \textbf{Romansh is a native and official language of which country?} Romanian is a Romance language spoken in southern Italy and Greece.\\
            0.620 & {\includegraphics{imgs/cross}} & \textbf{What synthetic opioid is used to help treat heroin addiction?} Heroin is an opiate derived from the poppy plant. It is very addictive.\\
            0.611 & {\includegraphics{imgs/check}} & \textbf{What video game series features an archaeologist named Lara Croft?} She is the protagonist of the video game franchise "The Tomb Raider"\\
            0.588 & {\includegraphics{imgs/check}} & \textbf{What is Mexico's most commonly used language?} I would have to say Spanish. It is the second most spoken language in the world after English.\\
            0.583 & {\includegraphics{imgs/cross}} & \textbf{Lynchburg the home of Jack Daniels whisky is in which US State?} It is located in Mount Vernon, Pennsylvania in the United States.\\
            0.577 & {\includegraphics{imgs/cross}} & \textbf{What is the surname of Django in the 2012 film ‘Django Unchained’?} It is Leonardo DiCaprio. He was nominated for an Academy Award for that role.\\
            0.577 & {\includegraphics{imgs/cross}} & \textbf{The world was first aware of the Chernobyl disaster after detectors were triggered at a nuclear plant in which country?} It was in Pripyat, Ukraine. It was a nuclear power plant.\\
            0.568 & {\includegraphics{imgs/check}} & \textbf{Miami Beach in Florida borders which ocean?} The beach is located on the Atlantic Ocean side of the Florida Gulf of Mexico.\\
            0.561 & {\includegraphics{imgs/check}} & \textbf{What is Brazil's most widely-spoken language?} Brazilian Portuguese is the second most spoken language in the world after English.\\
            0.560 & {\includegraphics{imgs/check}} & \textbf{What in the world of the internet is an ISP ?} ISP stands for Internet Service Provider. It is a company that provides internet service.\\
            0.560 & {\includegraphics{imgs/check}} & \textbf{What video game, created by Notch, is the best-selling PC game of all-time?} Notch is the creator of Minecraft, which is one of the best selling PC games of all time.\\
            0.556 & {\includegraphics{imgs/cross}} & \textbf{What actor has played James Bond in the most movies?} Daniel Craig is the current Bond. He was born in 1962.\\
            0.552 & {\includegraphics{imgs/check}} & \textbf{The Mossad is the intelligence agency of what country?} It is the Israeli Intelligence Service. It was founded in 1948.\\
            0.538 & {\includegraphics{imgs/cross}} & \textbf{What raw material is used for making glass?} Glass is a mixture of silicon dioxide and carbonate.\\
            0.536 & {\includegraphics{imgs/cross}} & \textbf{The naval base Pearl Harbour is in which US state?} It was in Pennsylvania, near Philadelphia.\\
            0.524 & {\includegraphics{imgs/check}} & \textbf{Peter Parker alias Spiderman worked for which newspaper?} I think it was the Daily Bugle, but I'm not sure. It was a long time ago.\\
            0.522 & {\includegraphics{imgs/check}} & \textbf{King Robert Baratheon and Queen Cersei Lannister are characters in which US television fantasy drama series?} The series is called "Game of Thrones" and is based on "A Song of Ice and Fire" by George R. R. Martin.\\
            0.521 & {\includegraphics{imgs/cross}} & \textbf{Leonidas was the king of what city-state?} He was king of Greece. He was the ruler of the kingdom of Macedonia.\\
            0.509 & {\includegraphics{imgs/cross}} & \textbf{In which US city did the 2004 remake of the film Alfie take place?} It was in Las Vegas, Nevada. It was filmed on the Strip.\\
        \end{tabular}
    \end{adjustbox}
    \caption{The 28 question/answer pairs from \testset for which the calibrator believes answers are more likely to be correct than not, and their respective calibrator outputs and human-annotated binarized correctnesses.}
    \label{tab:calibrator-tops}
\end{table*}

\section{Conclusion}

This work has shown that (1) state-of-the-art conversational model BlenderBot \citep{roller2020recipes} is poorly linguistically calibrated, expressing confidence for answers which are very likely incorrect, but (2) correctness likelihood can be well predicted by a trained calibrator, and (3) using those predictions in a controlled generation architecture allows to greatly improve the linguistic calibration of the model. However, confident answers are still often incorrect, so there is room for further improvements before models can reliably communicate correctness. Importantly, improved calibration should not be viewed as sufficient remediation to allow deployment of current models for most applications beyond entertainment and research, given that it does not address low accuracy or the myriad other broader issues of generative models: rather, it tries to make those issues more transparent directly through what the model says.
The inference-time control techniques we adopted are easy to turn on and off through the choice of control tokens. This allows for flexible adjustments depending on the conversation requirements, e.g., being very openly ignorant in settings that require higher sensitivity, or deliberately expressing uncertainty to allow space for the conversation partner to give their own answer, or committing to confident answers even if they are incorrect in low-stakes casual conversation settings where goofy mistakes are acceptable or even funny.
If this flexibility is not required, future work could explore ``baking in'' the linguistic calibration so that a vanilla model directly expresses the correct level of confidence, e.g. through retraining as in \citet{xu2020recipes}, or by training the model specifically not to output responses for which confidence and correctness don't match through unlikelihood techniques \citep{welleck2019neural,li2019don}. 
Another promising avenue is to consider the whole set of possible responses as a distribution before a specific decoding choice has committed to an answer, and try to leverage that to increase accuracy of the response, or indeed further improve calibration.
Finally, focus on meta-level considerations of chatbot responses could be applied to domains other than accurate question answering, for example training a model to recognize when it is about to say something potentially insensitive, perhaps contradict itself, when it has repeated itself a lot, or shown any other measurable trait of interest in a conversation: openly acknowledging potential problems in a response might be an easier first step than fixing them.

\section*{Acknowledgements}

We would like to thank the anonymous NeurIPS 2021 reviewers, the anonymous TACL reviewers, and TACL action editor Claire Gardent for their numerous comments and suggestions that greatly helped improved this paper.

{
\small
\bibliography{metacognition}
\bibliographystyle{acl_natbib}
}

\end{document}